\documentclass[10pt,twocolumn,letterpaper]{article}

\usepackage{cvpr}
\usepackage{times}
\usepackage{epsfig}
\usepackage{graphicx}
\usepackage{amsmath}
\usepackage{amssymb}

\usepackage{amsmath}
\usepackage{amssymb}
\usepackage{booktabs}
\usepackage{array}
\usepackage{floatrow}
\usepackage[table,x11names]{xcolor}
\usepackage{soul}
\usepackage{bm}
\definecolor{LightCyan}{rgb}{0.88,1,1}
\DeclareMathOperator{\softmax}{softmax}

\DeclareMathOperator{\lstm}{LSTM}
\DeclareMathOperator{\relu}{ReLU}
\newcommand{\fnote}[1]{\textcolor[rgb]{0,0,0}{{#1}}}

\newcommand{\keypoint}[1]{\noindent\textbf{#1}\quad}

\usepackage[pagebackref=true,breaklinks=true,letterpaper=true,colorlinks,bookmarks=false]{hyperref}

\cvprfinalcopy % *** Uncomment this line for the final submission

\def\cvprPaperID{2044} % *** Enter the CVPR Paper ID here

\ifcvprfinal\fi
\begin{document}

%%%%%%%%% TITLE
\title{iVQA: Inverse Visual Question Answering}

\author{Feng Liu$^{1}$ \quad Tao Xiang$^2$ \quad Timothy M. Hospedales$^3$ \quad Wankou Yang$^1$ \quad Changyin Sun$^1$\\
$^1$Southeast University, China 
\\$^2$Queen Mary University of London, UK \quad  $^3$University of Edinburgh, UK \\
{\tt\small \{liufeng,wkyang,cysun\}@seu.edu.cn,\quad t.xiang@qmul.ac.uk,\quad t.hospedales@ed.ac.uk}
}

\maketitle
%\thispagestyle{empty}

%%%%%%%%% ABSTRACT
\begin{abstract}
\vspace{-0.3cm}
We propose the inverse problem of Visual question answering (iVQA), and explore its suitability as a benchmark for visuo-linguistic understanding. The iVQA task is to generate a question that corresponds to a given image and answer pair. Since the answers are less informative than the questions, and the questions have less learnable bias, an iVQA model needs to better understand the image to be successful than a VQA model. We pose question generation as a multi-modal dynamic inference process and propose an iVQA model that can gradually adjust its focus of attention guided by both a partially generated question and the answer. For evaluation, apart from existing linguistic metrics, we propose  a new ranking metric.
This metric compares the ground truth question's rank among a list of distractors, which allows the drawbacks of different algorithms and sources of error to be studied. Experimental results show that our model can generate diverse,  grammatically correct and content correlated questions that match the given answer. 
\end{abstract}

%%%%%%%%% BODY TEXT
\vspace{-0.5cm}
\section{Introduction}
% -------------- Motivation --------------------
As conventional object detection and recognition approach solved problems, we see a surge of interest in more challenging problems that should require greater `understanding' from computer vision systems. Image captioning \cite{showandtell}, visual question answering \cite{vqa_dataset}, natural language object retrieval \cite{KazemzadehOrdonezMattenBergEMNLP14} and `visual Turing tests' \cite{geman2015visualTuringTestVQA} provide multi-modal AI challenges that are expected to  require rich visual and linguistic understanding, as well as knowledge representation and reasoning capabilities. As interest in these grand challenges has grown, so has scrutiny of the benchmarks and models that appear to solve them. Are we making progress towards these  challenges, or are good results the latest incarnation of horses \cite{pfungst1991cleverHans,sturm2014horse} and Potemkin villages \cite{goodfellow2015adversarialExamples}, with neural networks finding unexpected correlates that provide shortcuts to give away the answer? 

% -------------- Related works and drawbacks ---
Recent analyses of VQA models and benchmarks have found that the reported VQA success is largely due to making  predictions from dataset biases and cues given away in the question, with predictions being minimally dependent on understanding image content. For example it turns out that existing VQA models do not `look' in the same places as humans do to answer the question \cite{vqahat}; they do not give different answers when the same question is asked of different images \cite{vqa_analysis1};  and they can  perform well given no image at all \cite{vqa_dataset,jabri2016revisitVQA}. Moreover, VQA model predictions do not depend on more than the first few words of the question \cite{vqa_analysis1}, and their success depends largely on being able to exploit label bias \cite{VQA2.0}. These observations have motivated renewed attempts to devise more rigorous VQA benchmarks \cite{VQA2.0}.

\begin{figure}[t]
\includegraphics[width=0.8\columnwidth]{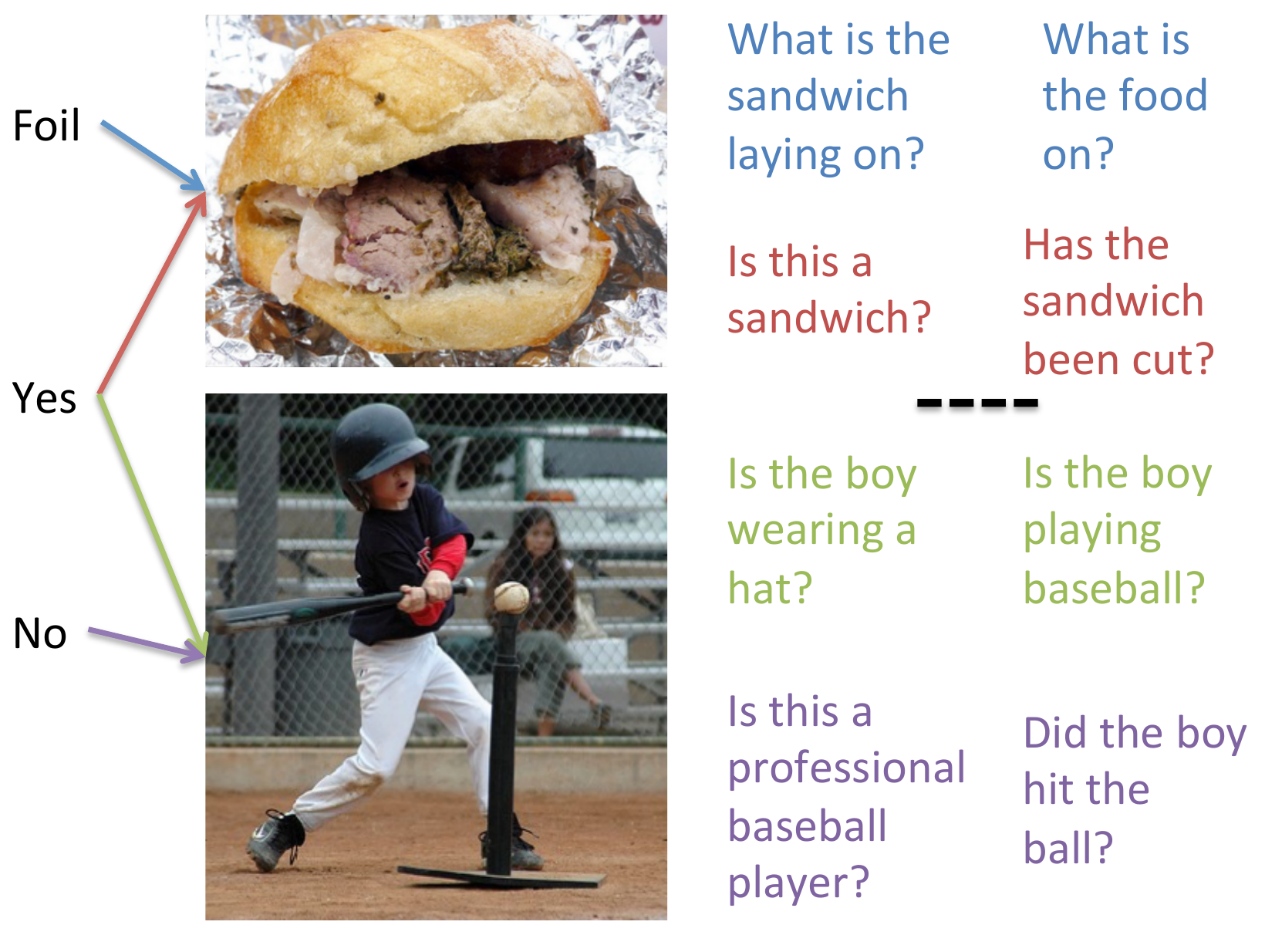}
\vspace{-0.3cm}
\caption{Illustration of iVQA task: Input answers and images along with the top questions generated by our model. }\label{fig:overview}
\end{figure}

%--- Our paper ---% 
In this paper we take a different approach, and explore whether the  task of inverse VQA provides an interesting benchmark of multi-modal intelligence. The inverse VQA (iVQA) task is to input a pair of image and answer, and then ask (output) a suitable question for which the given answer holds in the context of the given image. We conjecture that iVQA, as illustrated in Fig.~\ref{fig:overview}, is an interesting challenge for several reasons: (i) There may be less scope for an iVQA model to take advantage of question bias than for VQA to score highly through answer bias (there is less question bias, and exploiting it is harder than for categorical answers). (ii) The answers themselves provide a very sparse cue in iVQA compared to questions in VQA. So there may be less opportunity to deduce the question from the answer alone in iVQA than there is to deduce the answer from the question alone in VQA. Thus the iVQA task relies more heavily on understanding image content. (iii) From a knowledge representation and reasoning perspective, iVQA may provide the opportunity to test more complex inference strategies such as counterfactual reasoning \cite{bottou2013counterfactual}.
%TH: Do we have scope to say anything about working with one-to-many being more interesting/challenging? There is one right answer, but possibly many corresponding questions. Or downplay this because the dataset is not well organised for it, so can be seen as a flaw?

%----- The specific model proposed ---

Although closely related to VQA, existing VQA models do not provide a solution to the iVQA problem. This is because much less information can be inferred from an answer than from a question. In addition, although an answer is often short consisting of a phrase or even a single word, an iVQA model-generated question is a complete sentence composed of a long sequence of words. The key to effective iVQA is thus to attend selectively and dynamically to different regions of the image as the model progresses to generate the next word. This dynamic attention mechanism has to be conditioned on both the answer and the partial sentence generated so far. To this end, a novel dynamic multi-modal attention-based iVQA model is proposed which is capable of generating diverse, grammatically correct and content correlated questions that match the given answer.

% ------------- The Evaluation method ------------
Prior evaluations of question generation methods mainly use standard machine translation metrics, \eg, BLEU, METEOR,  \etc. These automatic metrics are correlated with human judgements for question generation \cite{vqg}. %, and they enable evaluating unconstrained  open-ended sentence generation.  
However they provide limited power to \emph{diagnose} question generation models in terms of when and why they succeed or fail. In this paper, we first propose an alternative and complementary ranking-based evaluation metric which is based on ranking the ground truth question among alternative distractors using an iVQA model, given the image and the answer.
%, where the rank of different questions is computed according to their likelihood conditioned on the image and answer. 
By controlling the types of distractors presented when using this metric, we can better understand the successes and failures of different models. Second, we perform a human study which is robust to iVQA's one-to-many nature (multiple possible questions can have the same answer). Reassuringly our human study scores turn out to be highly correlated to our proposed new ranking metric.

% -------------- Contribition ------------------
The contributions of this paper are as follows: (1) The novel iVQA problem is introduced as an alternative challenge for high-level multi-modal visuo-linguistic understanding. (2) We propose a multi-modal dynamic attention based iVQA model. (3) We propose a question ranking based evaluation methodology for iVQA that is helpful to diagnose the strengths and weaknesses of different models. (4) As the dual problem of VQA, we show that iVQA has the potential to help improve VQA performance.

\vspace{-0.3cm}
\section{Related work}
\vspace{-0.1cm}
\vspace{0.1cm}\noindent\textbf{Image captioning}\quad Image captioning \cite{showandtell,feifeicap} aims to describe, rather than merely recognise objects in images. It encompasses a number of classic vision capabilities as prerequisites including object \cite{resnet} and action  \cite{actionRecognise} recognition, attribute description \cite{farhadi2009attrib_describe} and relationship inference \cite{lu2016visualRelationshipLanguage}. It further requires natural language generation capabilities to synthesise open-ended linguistic descriptions. Popular benchmarks and competitions have inspired intensive research in this area. Captioning models have explicitly addressed these sub-tasks to varying degrees \cite{feifeicap}, but the most common and successful approaches use neural encoders (of  images), and decoders (of  captions), with little explicit knowledge representation and reasoning \cite{showandtell,liu2017semanticRecurrent}. The iVQA task investigated here is related to captioning in that we aim to produce natural language outputs, but distinct in that the outputs are sharply conditioned on the required answer, as illustrated in Fig.~\ref{fig:overview}.

\vspace{0.1cm}\noindent\textbf{VQA Challenge}\quad Like captioning, VQA has gained attention as a synthesis challenge in AI, requiring both computer vision and natural language understanding to succeed \cite{vqa_dataset}. Based on an image, and a natural language question about the image, a VQA system  produces an answer. Unlike other vision tasks (recognition, detection, description), the question to be answered in VQA is dynamically specified at runtime. Besides visuo-linguistic grounding, many VQA examples seem to require extra information not contained in the question or image, \eg,  background common sense about the world. Thus VQA is hoped to provide a long term goal for  AI-complete multi-modal intelligence.  However increasing scrutiny has shown that learning systems excel at finding shortcuts in terms of gaming the biases in answer distributions, and giveaway correlations \cite{vqa_analysis1,jabri2016revisitVQA}, leading to doubts about the level of visuo-linguistic intelligence implied by current results \cite{vqahat}. Although some benchmarks in principle require open-sentence answers, most answers are simple one-word outputs, and therefore the most common approach has been to formalise answer generation as a multi-class classification problem over the most frequent answers \cite{jabri2016revisitVQA,mcb}. Although successful, this is somewhat unsatisfactory as it is no longer an open-world challenge. In this paper we explore a novel iVQA task as an alternative open-world benchmark for visuo-linguistic understanding. 

\vspace{0.1cm}\noindent\textbf{VQA Models}\quad Existing VQA models are commonly based on two-branch neural networks, each consisting of a CNN image encoder, a LSTM question encoder which are merged before feeding to an answer decoder \cite{vqa_dataset,malinowski2015askYourNeurons}. Recently they have been enhanced through various mechanisms including better visuo-linguistic merging \cite{mlb,mcb}, varying degrees of explicit representation \cite{andreas2016neuralModuleNet}, reasoning with external knowledge bases \cite{wang2016fvqa}, and improving  visual encoding through attention \cite{mcb,xu2016vqaAttention}. With most recent models treating answer generation as a classification problem, these models cannot be directly modified for iVQA by simply swapping the answer and question encoder/decoder. The proposed iVQA model is a marriage between captioning and VQA models but with a dynamic and multi-modal attention mechanism developed specificaly for iVQA.  %In generating unconstrained answers, iVQA provides an open-world challenge. \doublecheck{An interesting unique aspect of iVQA is that it is one to many. There are often multiple valid questions (outputs) that have to the same answer (input), unlike in VQA. By sampling from our iVQA model, we can explore the spread of questions the model believes to be correct (Fig.~\ref{fig:overview})}

\vspace{0.1cm}\noindent\textbf{Related Challenges}\quad %---- Discussion vs VQG -----
Our proposed challenge is related to emerging task of \textit{visual question generation} (VQG): to generate a natural question about the content of an image \cite{vqg}. Introduced in \cite{vqg}, VQG is further studied in  \cite{arxivVQG} where  DenseCap \cite{densecap} is used to generate region specific descriptions before being  translated into questions. VQG  is a pre-specified task unlike VQA and iVQA which are dynamically determined at runtime. Importantly, VQG is  easier in terms of required understanding. Since VQG is not required to be answer-conditional, it often generates very general open questions that even humans cannot answer. It does not need to understand the image clearly enough -- and ground the two domains richly enough -- to correctly condition the generated question on the answer.  Another relevant challenge is \textit{visually grounded conversation} (VGC) which  aims to generate natural-sounding conversations \cite{das2016visual,vqg_extend}. VGC typically starts with VQG, but the following responses and further questions are generated primarily following conversational patterns mined from social media text, and only loosely grounded on the context of image content. In contrast, the image content grounding is much tighter in iVQA.
%Some approaches to VQG have have leveraged existing tools by applying caption generation followed by using language processing techniques to convert text into questions \cite{arxivVQG}. We aim to achieve end-to-end iVQA.
%The visually grounded conversation aims to generate natural-sounding conversations}

\vspace{-0.1cm}
\section{Methodology}
\begin{figure}
\begin{center}
\includegraphics[width=0.8\textwidth]{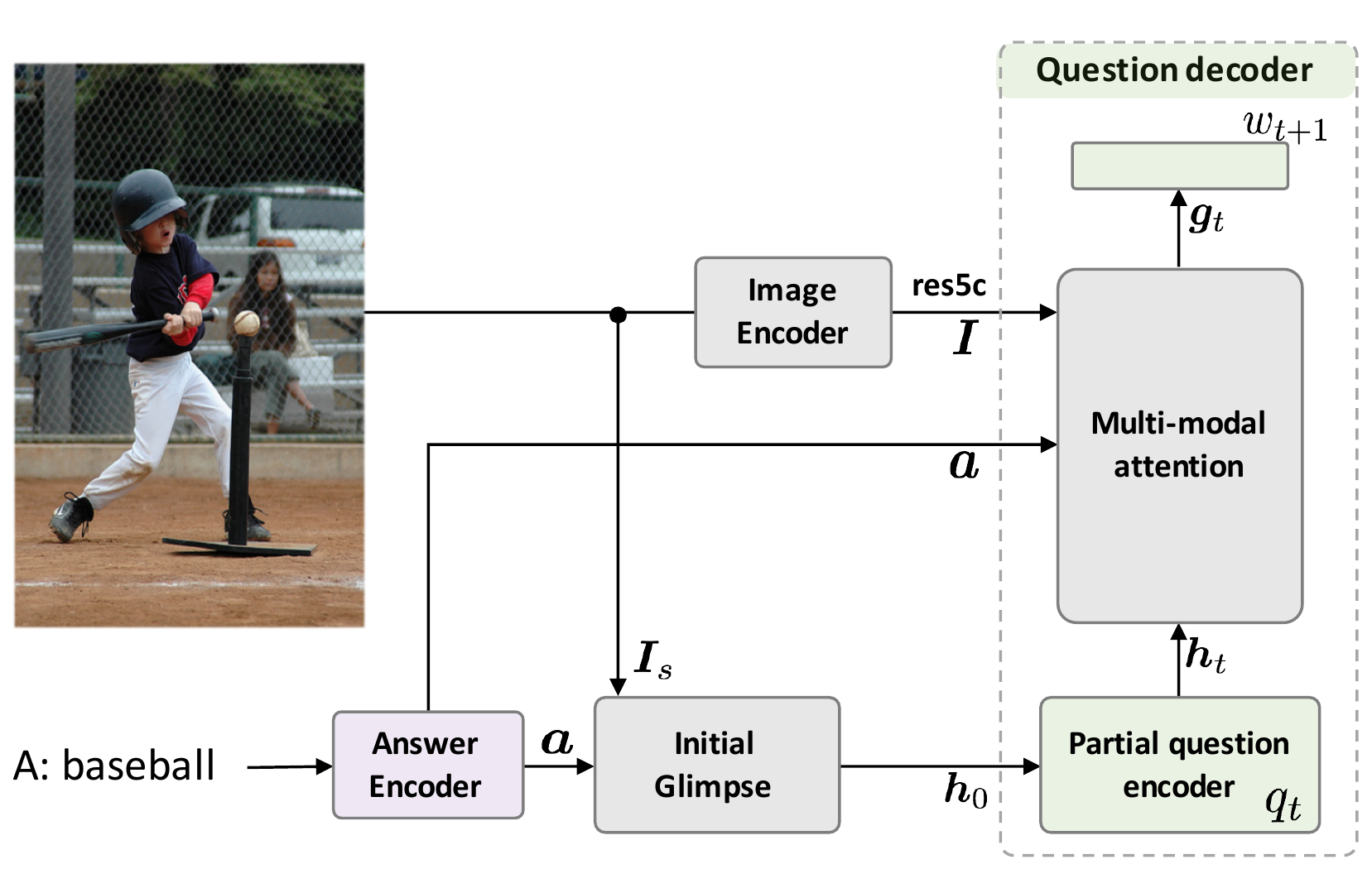}
\vspace{-0.5cm}
\caption{Overall architecture of the proposed iVQA model}\label{fig:overall}
\end{center}
%\vspace{-2mm}
\end{figure}

\subsection{Problem formulation}
The problem of inverse visual question answering (iVQA) is to infer a question $q$ for which a given answer $a$ holds, in the context of a particular image $I$. Formally:
\begin{equation}
q^{*} = \max_{q} p(q|I, a;\Theta),
\end{equation}
where $q$ is a sentence with words $(w_1,w_2,...,w_n)$ and $\Theta$ is the model parameters. As a sequence generation problem, we can use a recurrent neural network  language model, to generate the sentence by maximising the likelihood:
\begin{equation}
q^{*} = \max_{q} \prod_{t} p(w_{t}|w_{t-1},...,w_{1},I,a).
\end{equation}
Since the task is conditioned on both image $I$ and answer $a$, the visual information has to be integrated with the answers appropriately to generate questions. 

\subsection{Model overview}
\vspace{-0.2cm}

The architecture of our iVQA model is shown in Fig.~\ref{fig:overall}. It is a deep neural network with three subnets: an image encoder, an answer encoder, and a question decoder. The two encoders provide inputs for the decoder to generate a sentence which fits to the conditioned answer and image content. A multi-modal attention module (detailed later) is also a key component that directs image attention dynamically given the outputs of both encoders and a partial question encoder. We first describe the three subnets.

The image encoder is a CNN that generates a feature representation of the image. Both global  and local features are exploited for image representation. The \verb|res5c| features computed using the ResNet-152 model \cite{resnet} are utilised as local features. More specifically, The local feature collection $ \bm{I}=\{\bm{v}_{ij}\}$ is defined as local feature $\bm{v}_{ij}\in \mathbb{R}^{2048}$  over all $14\times 14$ spatial locations.  To extract the local features, we resize the image to $448\times448$ before feeding it to the feature extractor as in \cite{mcb}. As for the global feature, the semantic concept \cite{liu2017semanticRecurrent} feature $\bm{I}_s\in\mathbb{R}^{1000}$ is used. These 1,000 semantic concepts are mined from the most frequents words in a set of image captions. A concept classifier is learned to predict $\bm{I}_s$ as classification scores for the concepts.
 %feature \textcolor{magenta}{$I_g\in \mathbb{R}^{2048\times1}$ after global average pooling is} \textcolor{red}{$I_s$?} utilised as the global feature.

For the answer encoder, a long-short memory (LSTM) network with 512 cells \cite{lstm_cell} is used, and the concatenation of the final hidden state and cell state provides the answer representation $\bm{a} \in \mathbb{R}^{1024}$. 
With the described CNN image encoder and LSTM based answer encoder as input, a LSTM question decoder can be used to generate questions conditioned on both the images and answer. The detailed decoding processing together with the proposed attention module will be detailed next. %\doublecheck{Using the described CNN image encoder, and LSTM answer/question encoder/decoder, we have a complete potential iVQA system.} However, in the next section we discuss how to enhance this with a suitable attention model.

\vspace{-0.2cm}
\subsection{Dynamic multi-modal attention}\label{sub:attn}
\vspace{-0.2cm}
Given the sparse information contained in the answer, having an effective attention model to focus on the right region of the image  is critical for iVQA. Attention models have been widely studied in image captioning \cite{lu2017whenToLook} and VQA \cite{xu2016vqaAttention}. However, our iVQA problem has some unique characteristics, and thus needs a tailor-made attention module. Specifically,  compared with VQA,  iVQA requires multiple decoding steps and the focus of attention therefore needs to be dynamically changed accordingly. Also unlike image captioning, the generation process has multi-modal conditions: \ie, image and answer, both of which need to be integrated in every decoding step in a dynamic manner.   Consider the following question-answer pair: ``\emph{Q: What colour is the dress the girl is wearing?}''; ``\emph{A: Pink}''. Given the answer, the model can infer that the question is about colour. After the model has predicted type specific partial question $q_t=\{$\emph{what, colour, is, the}$\}$, the attention network will integrate the partial question $q_t$ with the answer $a=\{$\emph{pink}$\}$, and search for all objects with the pink colour and output attended features. Based on these attended features the next word $w_{t+1}$ is predicted.
Motivated by these unique characteristics, we propose an attention model that can perform inference based on image, partial question and answer jointly and dynamically. It is composed of the the following sub-modules:
% To effectively captures the details of the image and pursuit a diversified question generation result, we propose to simultaneously performing inference about the visual cue during the question generation process. It is achieved by the proposed attention mechanism, during each encoding step 

\vspace{0.1cm}\noindent\textbf{Initial glimpse}\quad The initial glimpse should provide an overview cue of the input image-answer pair, to establish a  good starting point for the decoding process. We use semantic concept prediction $\bm{I}_s$  as a global visual cue, which captures 1-gram information that may be relevant to the question \cite{liu2017semanticRecurrent}. The encoded answer $\bm{a}$ is taken as the textual cue, which determines the set of likely initial words of the target question. The two cues are integrated as %\todo{(Is vs Ig. Additive?)}
\begin{equation}
\label{eq:init_glimpse}
\bm{h}_{0} = \delta(\bm{W}_{ih} \bm{I}_s + \bm{W}_{ah} \bm{a}),
\end{equation}
where $\bm{W}_{ih}$ and $\bm{W}_{ah}$ are embedding weights\footnote{In all equations, we omit the bias term for simplicity.}, and $\delta(\cdot)$ is a tanh activation function. This joint representation is directly used as the initial memory of the decoding network.

%The decoding network is made up of several important sub-modules.

\vspace{0.1cm}\noindent\textbf{Encoding of partial question}\quad The partial question encoder sequentially encodes the partial question generated thus far \textcolor{black}{$q_t=\{w_1, w_2,...,w_t\}$} to a hidden representation $\bm{h}_t$. A LSTM network with 512 cells is used to encode the partial question to a hidden representation $\bm{h}_t$ as 
\begin{equation}
\begin{split}
& \bm{x}_t = \bm{E} \bm{w}_{t}, \\
& \bm{h}_t, \bm{m}_t = \lstm(\bm{x}_t, \bm{h}_{t-1}, \bm{m}_{t-1}), \\
\end{split}
\label{eq:partq_encode}
\end{equation}
where $\bm{w}_t$ is the one-hot coding of word $w_t$; $\bm{E}$ is the word embedding matrix; $\bm{x}_t$ is the embedded word vector which serves as an input to the LSTM.  $\lstm(\cdot)$ takes previous states $(\bm{h}_{t-1}, \bm{m}_{t-1})$ and $\bm{x}_t$ as input to generate the next states. For the computation of the LSTM, readers are referred to \cite{Hochreiter:1997:LSTM} for  details. 

\vspace{0.1cm}\noindent\textbf{Multi-modal attention network}\quad The attention network \textcolor{black}{takes local features $\bm{I}$}, partial question coding $\bm{h}_t$, and answer coding $\bm{a}$ as input, and outputs the joint embedding of attended visual features $\bm{c}_t$ specified by the partial question-answer context $\bm{z}_t$. To obtain the partial question-answer context, the partial question coding $\bm{h}_t$ and answer coding $\bm{a}$ are fused as
\begin{equation}
\label{eq:qa_context}
\bm{z}_t = \relu(\bm{W}_q  \bm{h}_{t} +  \bm{W}_a \bm{a}) \\
\end{equation}
where $\bm{W}_q$ and $\bm{W}_a$ are the embedding weights for question and answer respectively. 

Then the visual features $\bm{I}$ and textual context $\bm{z}_t$ are spatially matched via soft attention: The visual features \textcolor{black}{$\bm{v}_{i,j}$} and context vector $\bm{z}_t$ are fused by a multi-modal low rank bilinear pooling (MLB) \cite{mlb}, and then the fused feature $\bm{f}_{ij}$ is used for attention map computation as follows:
\begin{equation}
\label{eq:soft_attention}
\begin{split}
& \bm{f}_{ij} = \delta(\bm{U}\delta(\bm{W}_v  \bm{v}_{i,j}) \odot  \delta(\bm{W}_z \bm{z}_t)) \\
& \alpha_{ij}^t = \softmax(\bm{p}^{T}[\bm{f}_{ij}]) \\
& \bm{c}_t = \sum_{ij} \alpha_{ij}^t \bm{v}_{ij},
\end{split}
\end{equation}
where  $\bm{c}_t$ is the attended visual feature; $\bm{\alpha}_t =[\alpha_{ij}^t]$ is the attention map; and $\bm{U}$, $\bm{W}_v$, $\bm{W}_z$, $\bm{p}$ are the corresponding embedding weights.

The attended visual feature $\bm{c}_t$ is further fused with the textual context $\bm{z}_t$ via MLB, which can be interpreted as co-attention between vision and language \cite{zhou2016image}.
\begin{equation}
\bm{g}_t = \delta(\bm{U'}\delta(\bm{W}_c  \bm{c}_{t}) \odot  \delta(\bm{W}'_z \bm{z}_t)),
\end{equation}
where $\bm{U'}, \bm{W}_c, \bm{W}'_z$ are embedding weights of the pooling.

\vspace{0.1cm}\noindent\textbf{Word predictor}\quad The next-word predictor is a softmax classifier, which generates a distribution over the next words,  leveraging the multi-modal attention network's output $\bm{g}_t$:
\begin{equation}
w_{t+1} \sim \softmax(\bm{W}_{o} \bm{g}_t),
\end{equation}
where $\bm{W}_o$ are the classifier weights. The next word $w_{t+1}$ is sampled from the softmax classifier's distribution.

\vspace{-0.1cm}
\section{iVQA Evaluation}
We explore three of iVQA evaluation metrics including standard language-generation metrics, a new ranking-based metric, and a human validation study.

\vspace{0.1cm}\noindent\textbf{Linguistic Metrics}\quad
Standard linguistic measures \cite{DBLP:journals/corr/ChenFLVGDZ15} including  CIDEr, BLEU, METEOR and ROGUE-L  can be used to evaluate the generated questions. Given an image, question, answer tuple, we use the ground-truth question as the reference sentence, and compare the generated question based on the given image and answer. The similarity between the machine generated questions and the reference questions can be measured by these metrics. Even though generating humanlike questions is relatively easy, doing so in a correct image+answer conditional way  to get a high score is challenging, since the model has to capture all the semantic concepts and high-order interactions. 

\vspace{0.1cm}\noindent\textbf{Ranking Metric}\quad
We also develop a ranking  based evaluation metric for the iVQA problem. For an image-answer pair $(I,a)$, and a candidate question $q$, the conditioning score $p(q|I,a;\Theta)$ is used for ranking. If one of the correct (ground truth) questions is ranked the highest then this image-answer pair is regard as correct at Rank-1. In this way, accuracy over a testing set can be computed as the percentage of the times that correct questions are ranked at the top (denoted Acc.@1). Similarly, we can measure cumulative ranking accuracy at other ranks, e.g.,  Acc.@3 measures the  percentage of times  correct questions are ranked in top 3.  % In our experiments, 24 questions is used as candidates for evaluation. 
This is related to the 
%This is not open-world question answering, but more like the 
multiple-choice setting of VQA \cite{vqa_dataset}. However, the difference is that we can explore specific choices of candidate question subsets to form a question pool in order  to reveal insights about model strengths and weaknesses.

\vspace{0.1cm}\noindent\textbf{Question Pool}\quad
For a particular image-answer pair, the candidate ranking questions are collected from the following subsets. \textbf{Correct questions (GT)}: given image-answer pair $(I,a)$, the correct (ground truth) questions are defined as all the questions with answer $a$ in image $I$\footnote{There can be multiple correct questions corresponding to a given answer, since multiple questions may have the same answer for one image.}. \textbf{Contrastive questions (CT)}: these are questions associated with visually similar images to $I$ (including $I$) but having different answers. The similarity of the images is measured using the image CNN feature. \textbf{Plausible questions (PS):} These test whether the model can tell the subtle difference between questions and maintain grammar correctness. They are obtained by randomly replacing one of the key words (e.g., verbs, nouns, adjective, and adverb) in the ground truth question. \textbf{Popular questions (PP):} Popular questions are chosen to be the most popular questions with the same answer type as $a$ across the whole dataset. These diagnose the extent to which the model is relying on label-bias. \textbf{Answer-related (RN):} These are chosen to be the random questions having answer $a$ but from other images. These diagnose the extent to which the model is relying on visual features, which did not always happen in VQA \cite{VQA2.0}. We manually checked all generated distractor questions, and removed any which were also correct for their corresponding image answer.

\vspace{0.1cm}\keypoint{Human study} iVQA is open-ended and one-to-many in that there can be many correct questions for one image and answer. Therefore, given an image and an answer, the correct questions may not be annotated exhaustively in existing datasets. The proposed ranking metric computed on the selected question pool  ameliorates the effects of the open-ended/one-to-many questions generated by an iVQA model. However, it does not measure directly  how `correct' the generated questions are, when they differ from the human annotations originally provided.   To evaluate iVQA in a way that awards `credit' for correct questions that are not annotated originally, we perform a human evaluation study. Image-answer pairs are randomly selected from the test set, and annotators assess the generated questions, scoring them from 0 (complete nonsense) to 4 (perfect). The mean score is used as the metric.

\section{Experiments}
\subsection{Datasets and settings}
\keypoint{Dataset:} We repurpose the VQA dataset \cite{vqa_dataset} to investigate the iVQA task. The VQA dataset uses images from  MS COCO \cite{DBLP:journals/corr/ChenFLVGDZ15}, including 82,783 training, 40,504 validation  and 81,434 test images. Three question-answer pairs are collected for each image. Since the test set answers are not available, we adopt the commonly used off-line data split  in \cite{feifeicap,lu2017whenToLook} for image captioning: 82,783 images are used for training, and 5,000 for validation and test each. 

\vspace{0.1cm}\keypoint{Training:} The model is trained by minimising the cross entropy loss between machine generated questions and ground truth questions. The Adam \cite{adam} optimiser is employed with a batch size 32 for 30 epochs. The initial learning rate is set to be 5e-4, and it is annealed 0.83 times per epoch with an exponential decay.

\vspace{0.1cm}\keypoint{Inference:} Beam search with max sentence length 20 is used to generate questions. We use beam size 3 for quantitative and 10 for qualitative results.

\vspace{0.1cm}\keypoint{Question Pool:} For the proposed ranking accuracy metric, given each image-answer pair, the question pool  contains 24 questions, of which 1-3 are GT, 3-5 are CT (so that the total of GT+CT is 6), 6 are PP,  6 are PS, and 6 are RN .

\vspace{-0.2cm}
\subsection{Baseline models}
\vspace{-0.2cm}
\keypoint{Answer~only~(A):} It uses a LSTM encoder to encode tokenised answers to a fixed 512-dimensional representation, then a LSTM decoder to generate questions. 

\keypoint{Image only (I, VQG):} The visual only model is similar  to the GRNN model in \cite{vqg}, however, we use a more powerful image feature: the same \verb|res5c| feature of ResNet-152 \cite{resnet} used in our model. This feature is fed into a LSTM decoder as the initial state. 

\keypoint{Image+Answer Type (I+AT):}  VQG models \cite{vqg} generate questions purely based on visual cues. To make VQG more competitive in our answer-conditional iVQA setting, we also provide one-hot encoding of the answer \emph{type}. This hint helps a VQG model generate the right question type (e.g.,`is...', `what...'). 

\keypoint{NN:} We adapt the nearest neighbour (NN) image captioning method \cite{nn}  to our problem. As iVQA is conditioned on both image and answer, we averaged the distance computed from both modalities for NN computation.

\keypoint{SAT:} Show attend and tell (SAT) \cite{sat} is a strong attentional captioning method. To provide a strong competitor to our approach, we modify  SAT to take input from both modalities by setting  the initial state of the decoding LSTM as the joint embedding of image and answer.

\keypoint{VQG+VQA:} The VQG+VQA baseline uses the VQG model above to generate question proposals from the image, and then uses VQA to select the question with maximum conditioning score. We use VQG to generate 10 candidates for each image for VQA re-ranking, and the retrained multi-modal low-rank bilinear attention network \cite{mlb} is used as the VQA model.\\
\keypoint{Ours:} Our model processes images with local and global semantic features, and dynamic multi-modal attention (I+A+Att+$I_s$). The global semantic feature is obtained following \cite{liu2017semanticRecurrent} by learning a concept predictor on the training split using the 1,000 most frequent caption words.

\begin{figure*}[t]
\includegraphics[width=0.95\textwidth]{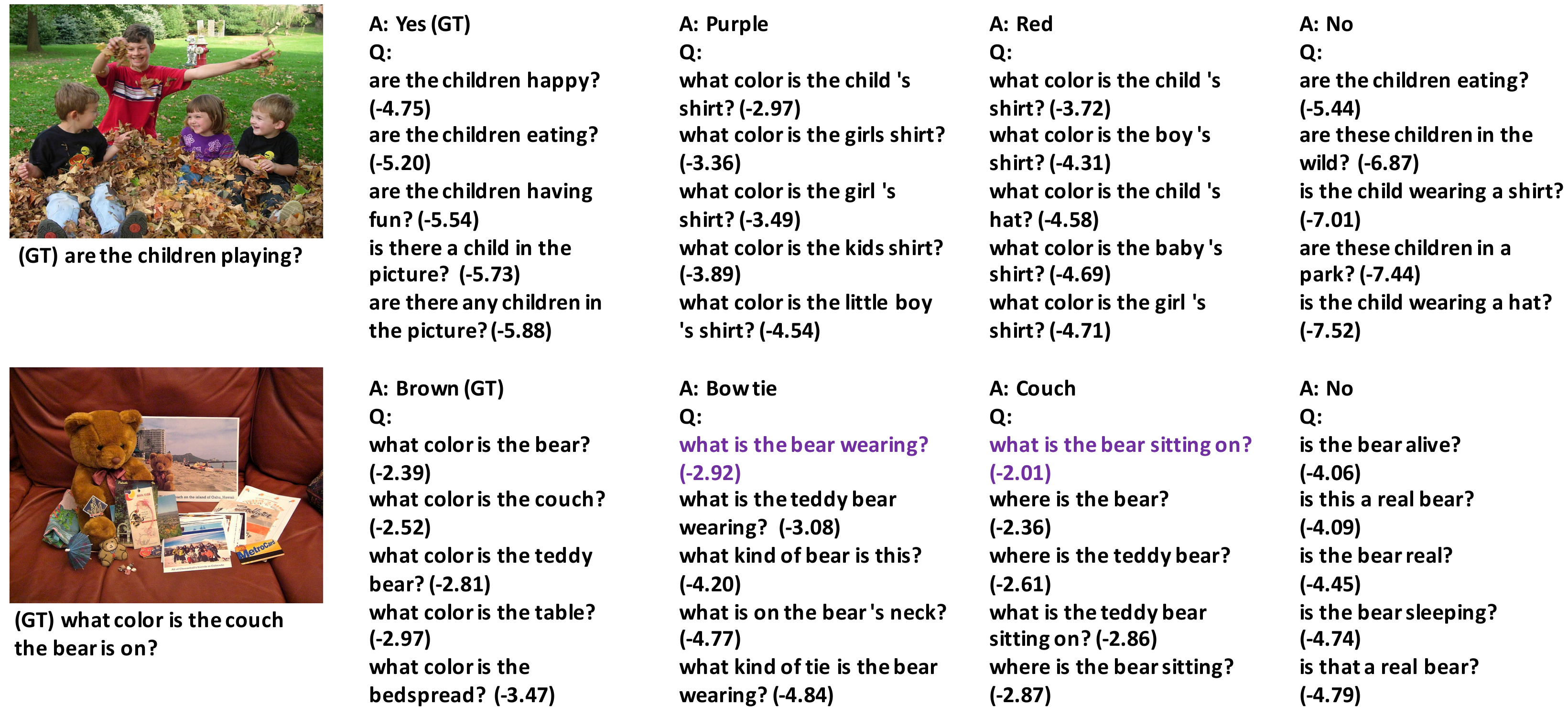}
\caption{Qualitative results of iVQA. Larger numbers in brackets mean higher confidence. The attention of generating the questions in purple is further visualised in Fig. \ref{fig:att_bear}.}\label{fig:qualitative}
\vspace{-0.1cm}
\end{figure*}

\begin{table*}[!thb]
\footnotesize
\begin{center}
\begin{tabular}{@{} l|c|cccc|c|c|cc|c @{}}
\toprule
 	& CIDEr	& BLEU-4	& BLEU-3	& BLEU-2	& BLEU-1	& ROUGE-L	& METEOR	& Acc@1	& Acc@3 & Human\\ 
\midrule 
A	& 0.952	& 0.146	& 0.192	& 0.265	& 0.371	& 0.408	& 0.161	& 14.589	& 28.795 & 1.92 \\ 
I	& 0.652	& 0.086	& 0.121	& 0.179	& 0.280	& 0.310	& 0.117	& 13.012	& 28.644 & 2.04\\ 
I+AT	& 0.904	& 0.122	& 0.164	& 0.234	& 0.350	& 0.397	& 0.151	& 20.277	& 36.134 & 2.65\\ 
\midrule
NN	& 1.372	& 0.175	& 0.223	& 0.294	& 0.404	& 0.428	& 0.183	& 26.783	& 48.755 & 3.01\\
SAT	& 1.533	& 0.192	& 0.241	& 0.311	& 0.417	& 0.456	& 0.195	& 29.722	& 48.118 & 3.19\\  
VQG+VQA	& 1.110	& 0.147	& 0.193	& 0.261	& 0.371	& 0.396	& 0.165	& 16.529	& 41.655 & 2.79\\ 
\midrule
Ours	& {\bf 1.714}	& {\bf 0.208}	& {\bf 0.256}	& {\bf 0.326}	& {\bf 0.430}	& {\bf 0.468}	& {\bf 0.205}	& {\bf 32.899}	& {\bf 51.418} & {\bf 3.31}\\ 
\bottomrule
\end{tabular}
\end{center}
\vspace{-0.4cm}
\caption{Overall question generation performance on the testing set.}
\label{tab:main_table}
\end{table*}

\begin{figure*}[t]
\begin{center}
\begin{tabular}{@{}l@{}l@{}l@{}l@{}l@{}l@{}l}
\includegraphics[width=0.16\linewidth]{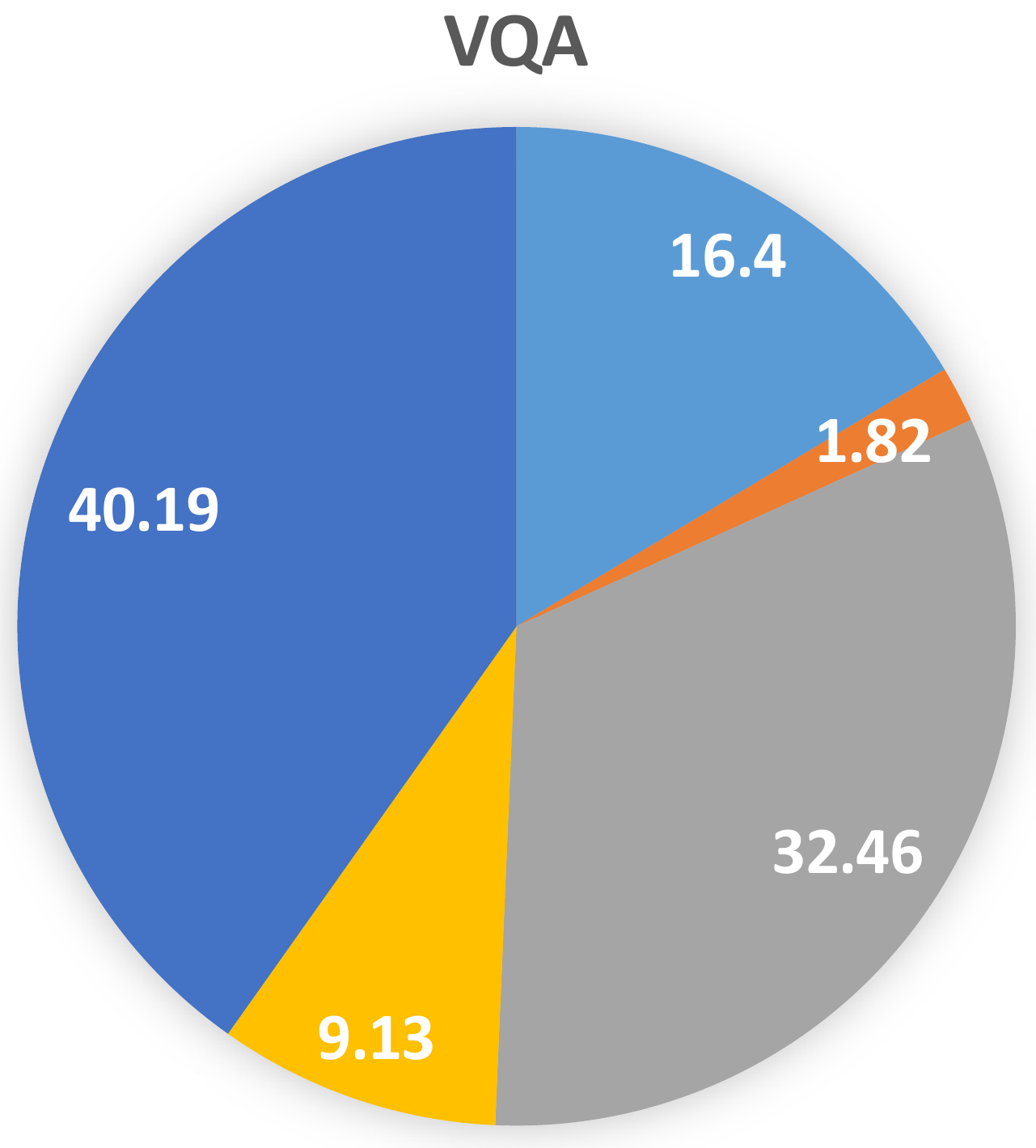}
& \includegraphics[width=0.16\linewidth]{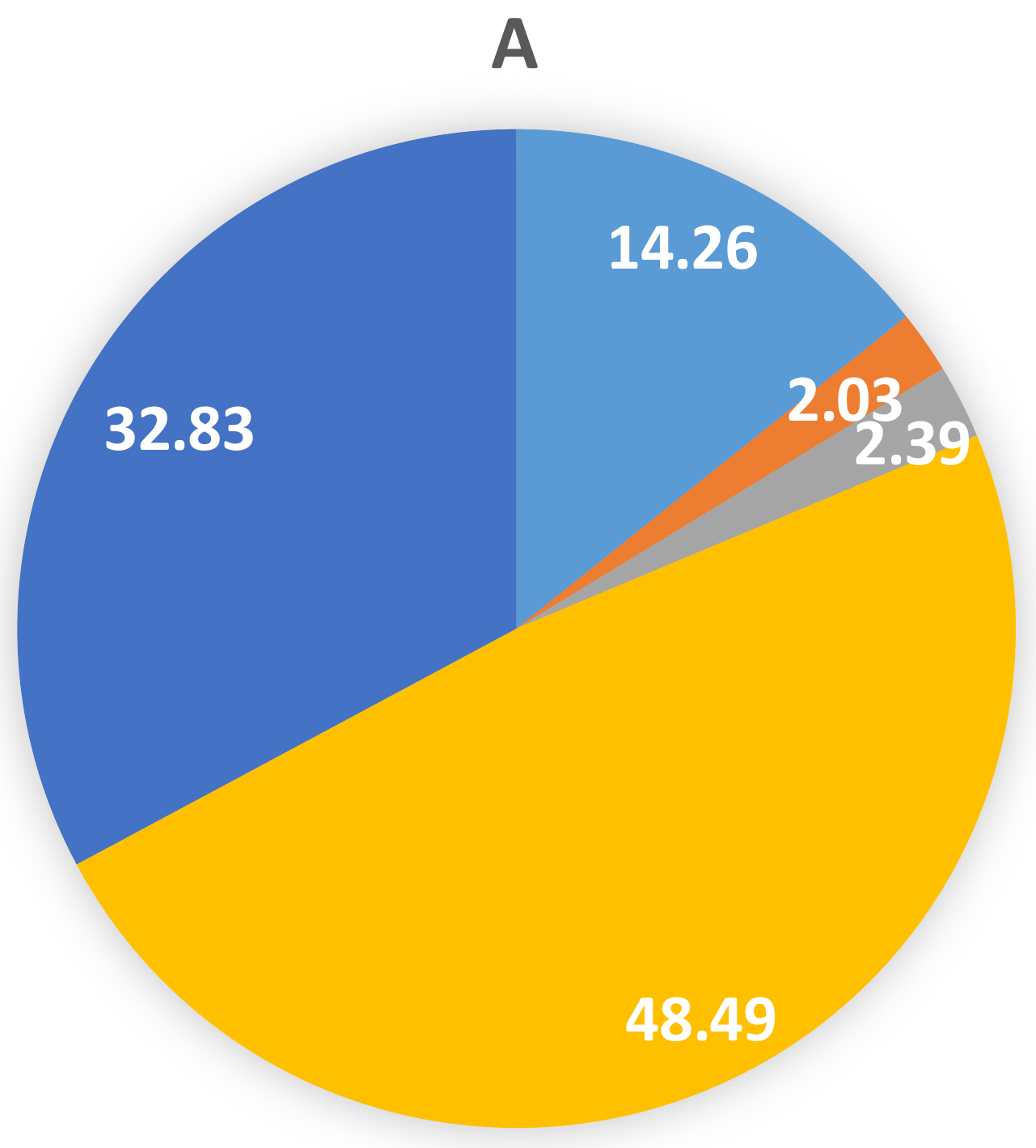}
& \includegraphics[width=0.16\linewidth]{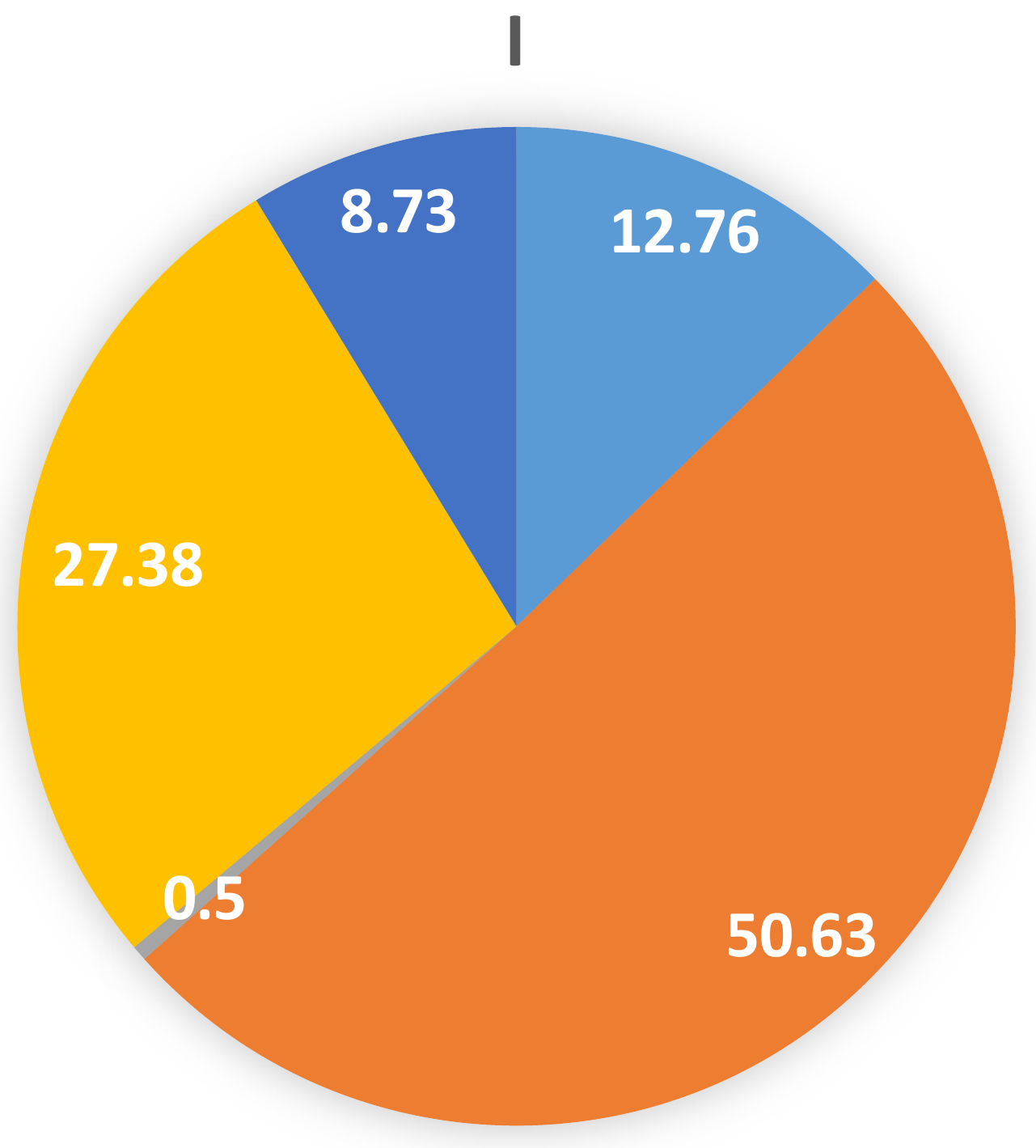}
& \includegraphics[width=0.16\linewidth]{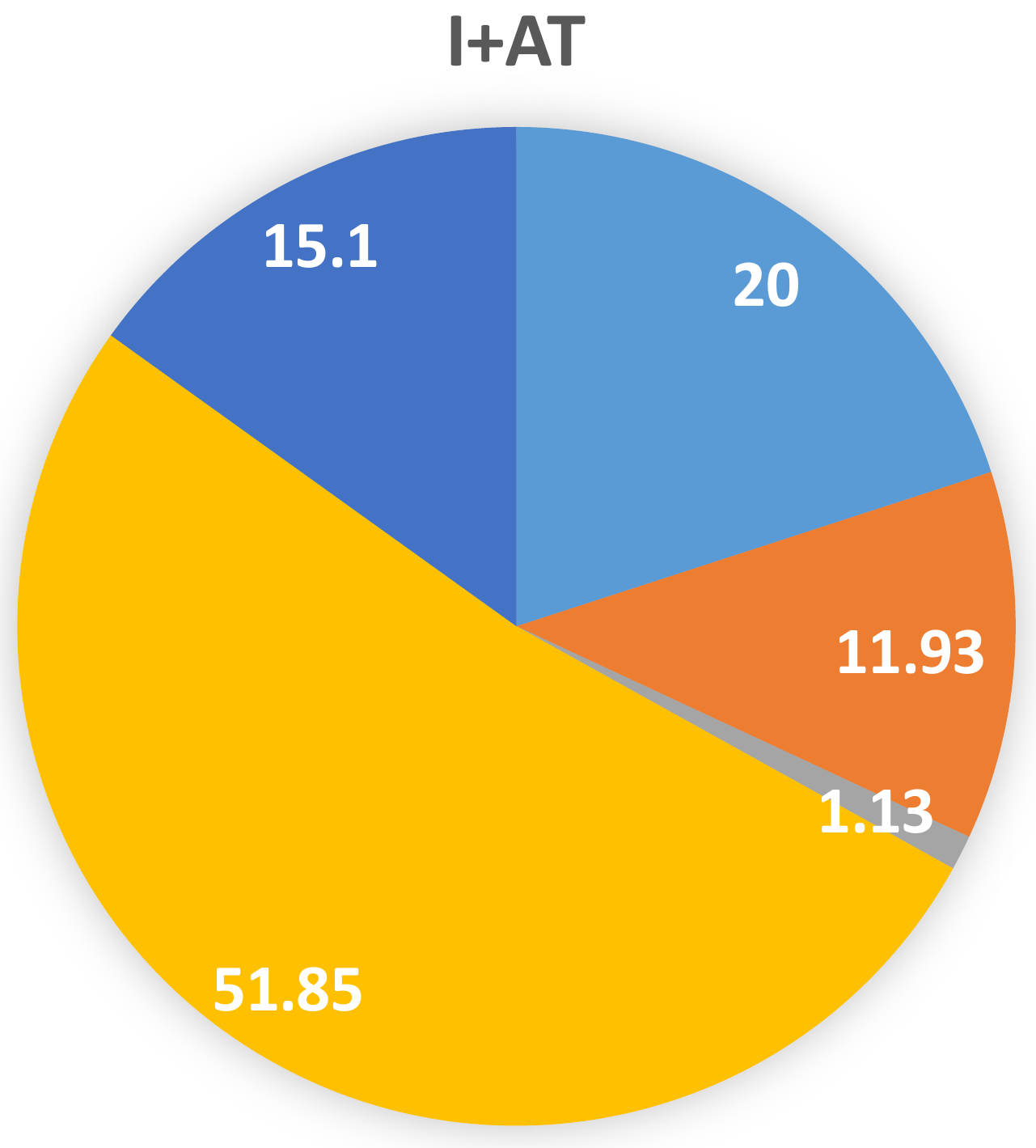}
& \includegraphics[width=0.16\linewidth]{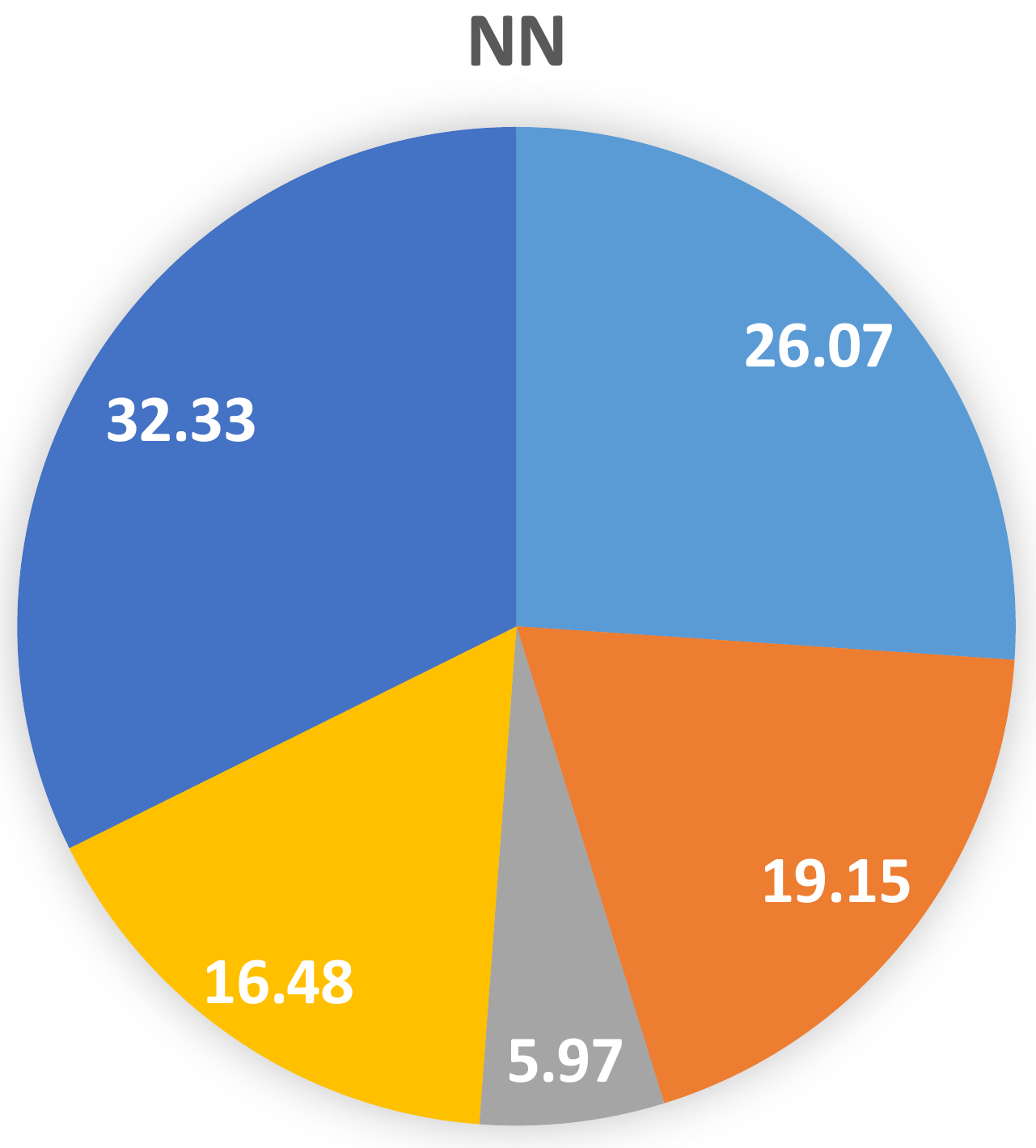}
& \includegraphics[width=0.16\linewidth]{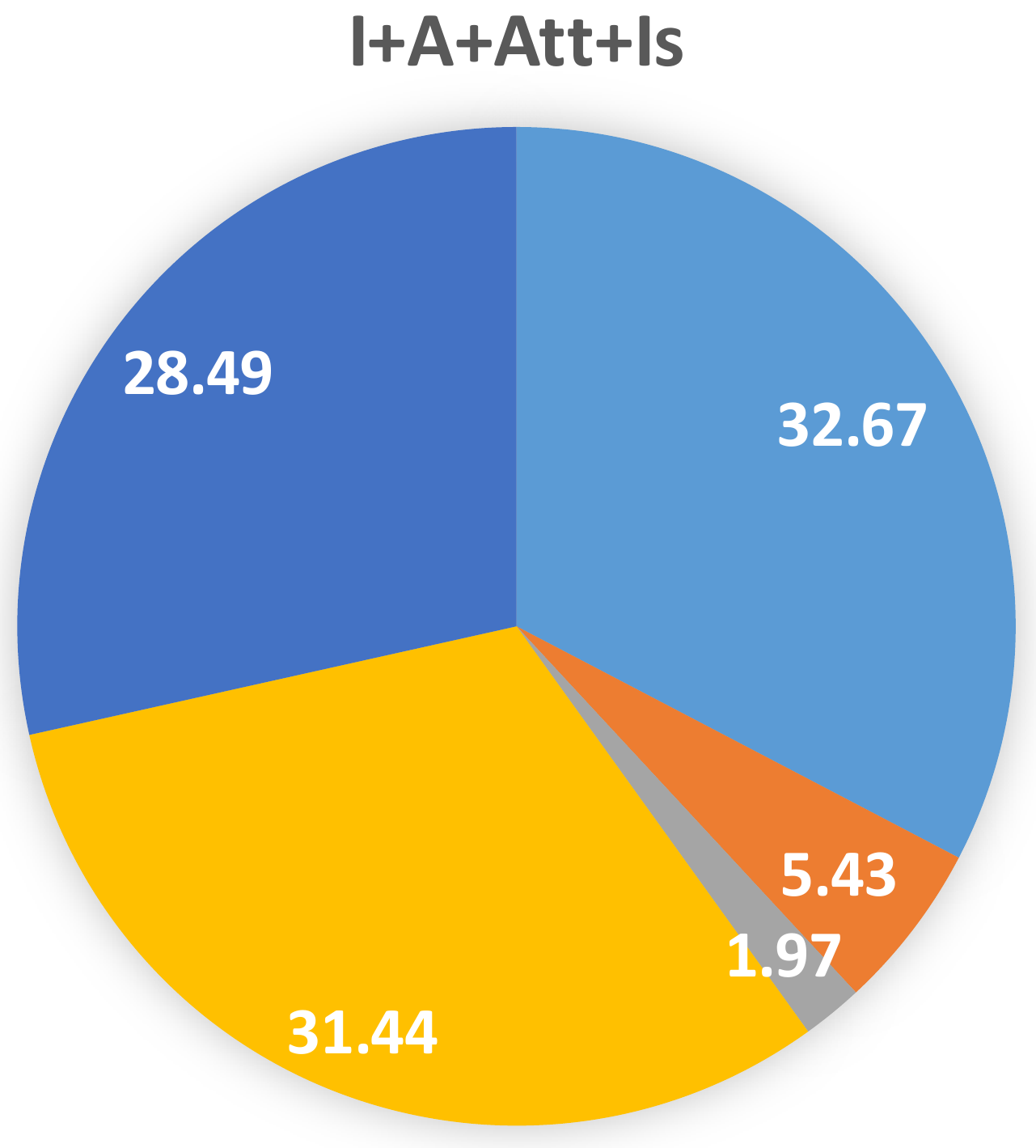}
& \ \includegraphics[width=0.05\linewidth]{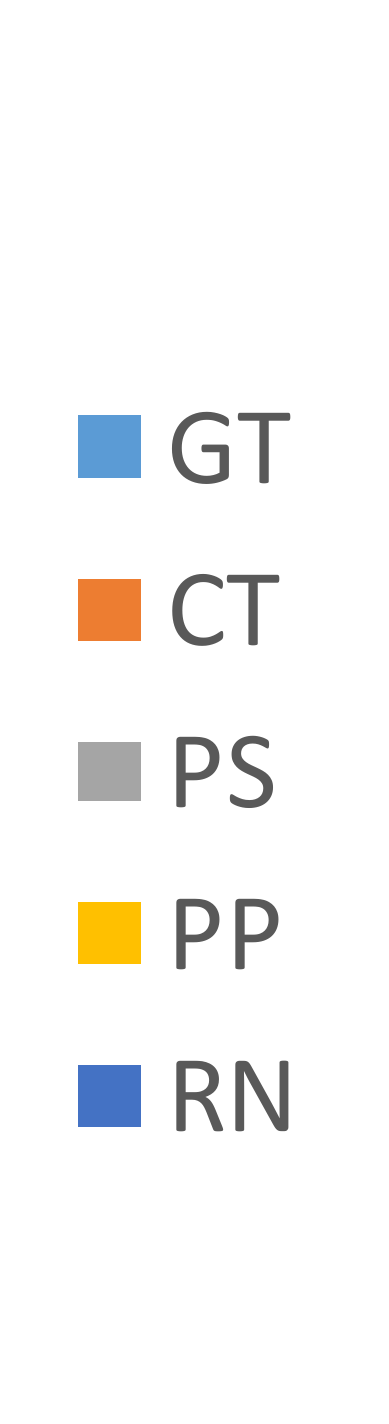}\\
\end{tabular}
\end{center}
\vspace{-0.6cm}
\caption{Comparison of effects of different distractors on different models on the test set.} \label{fig:distractorTypes}
\end{figure*}

% ******************* BEAR EXAMPLE ***************
\begin{figure*}[!ht]
\scriptsize
\begin{center}
%\begin{tabular}{@{}l@{}l@{}l@{}l@{}l@{}l@{}l@{}l@{}l@{}l}
\begin{tabular}{@{}c@{}c@{}c@{}c@{}c@{}c@{}c@{}c@{}}
\includegraphics[width=0.16\linewidth]{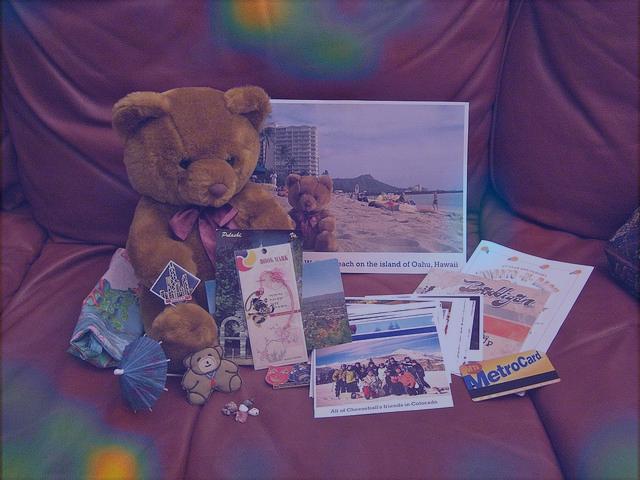}
&  \ \includegraphics[width=0.16\linewidth]{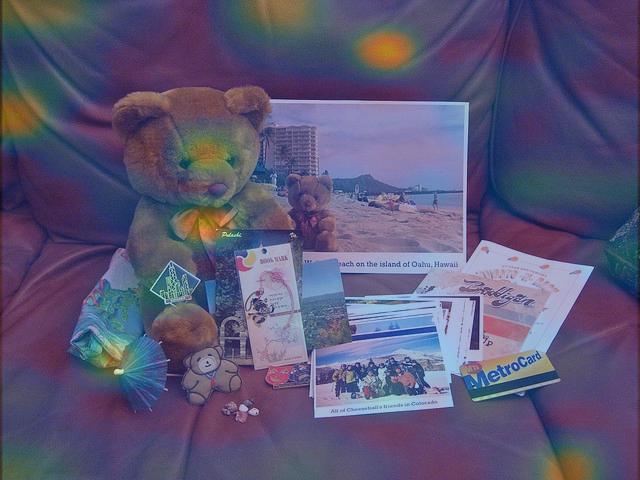}
&  \ \includegraphics[width=0.16\linewidth]{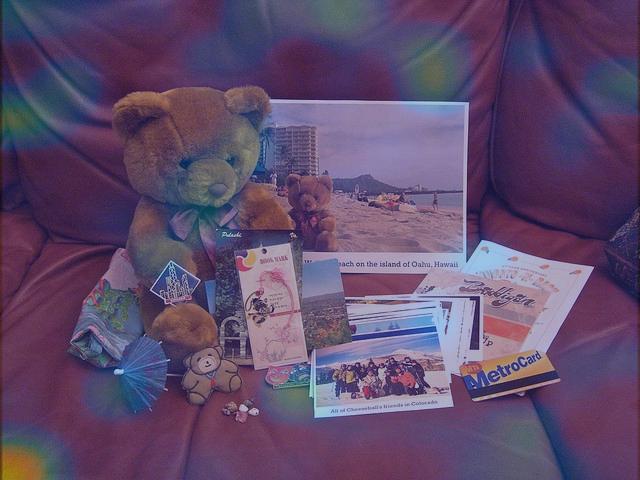}
&  \ \includegraphics[width=0.16\linewidth]{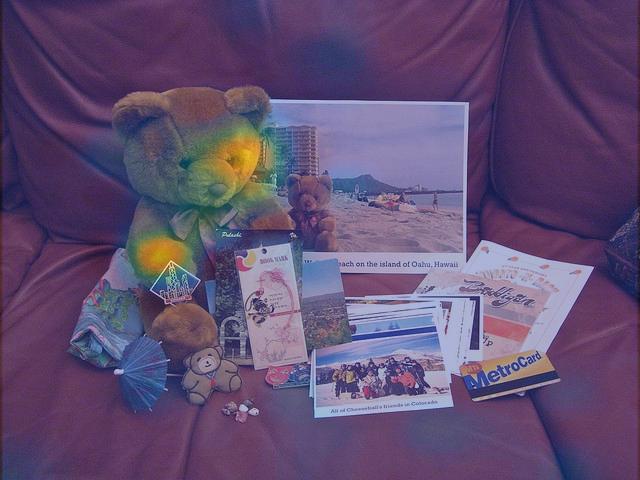}
&  \ \includegraphics[width=0.16\linewidth]{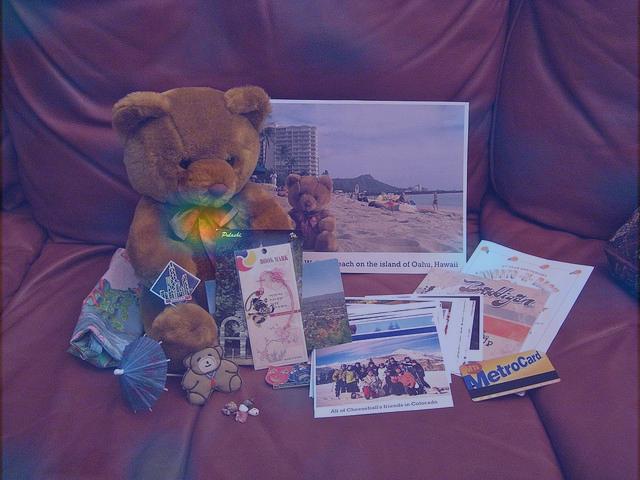}
&  \ \includegraphics[width=0.16\linewidth]{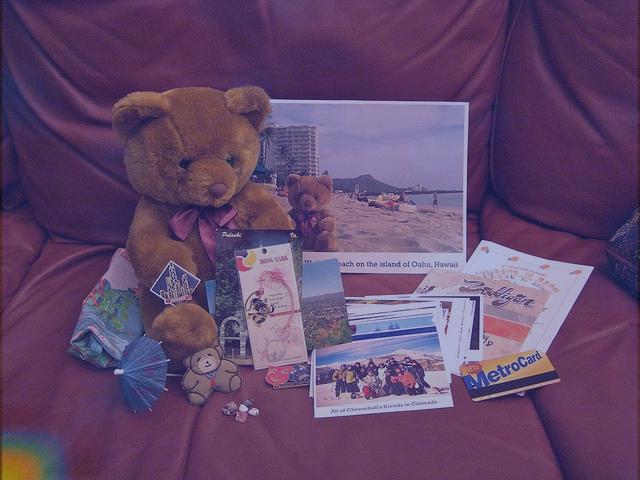} \\
% ------------- words ------------
what & is & the & bear & \textbf{wearing} & ? \\
% ---------------
\includegraphics[width=0.16\linewidth]{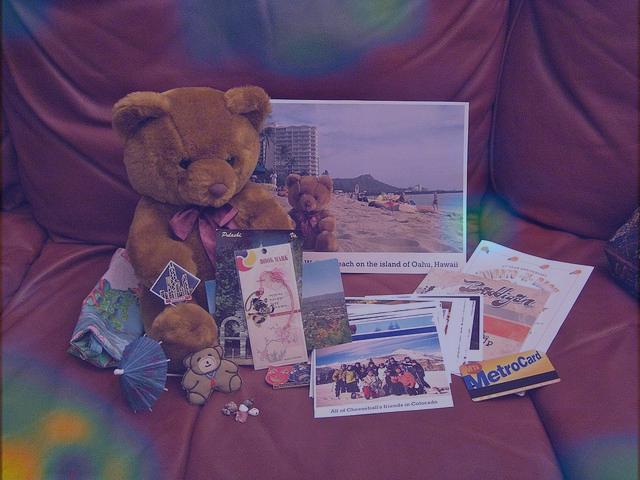}
&  \ \includegraphics[width=0.16\linewidth]{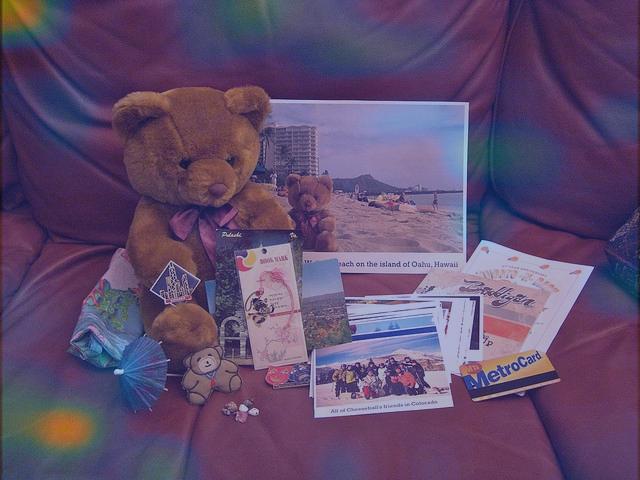}
&  \ \includegraphics[width=0.16\linewidth]{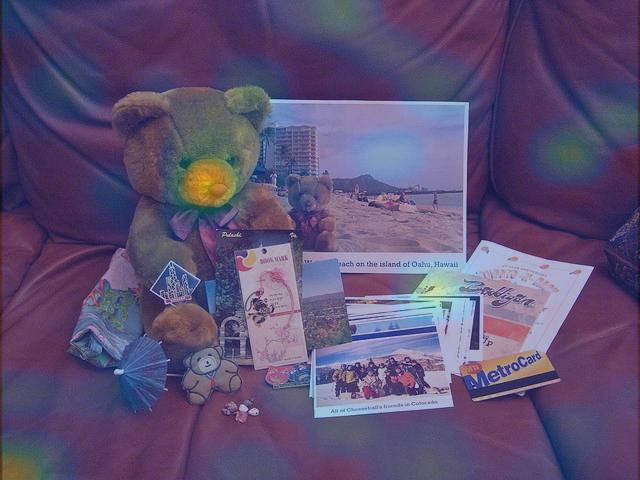}
&  \ \includegraphics[width=0.16\linewidth]{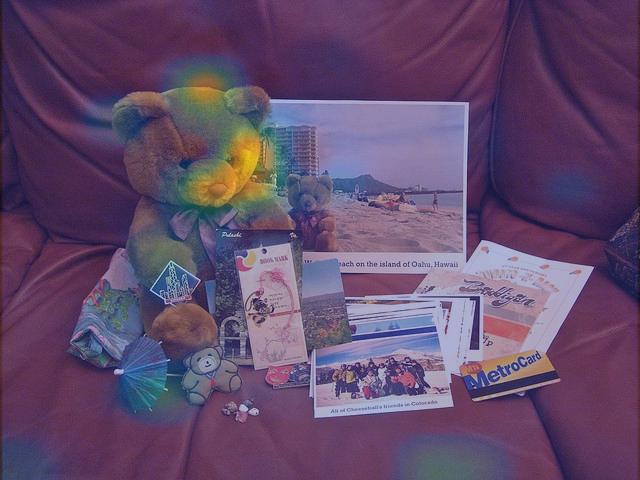}
&  \ \includegraphics[width=0.16\linewidth]{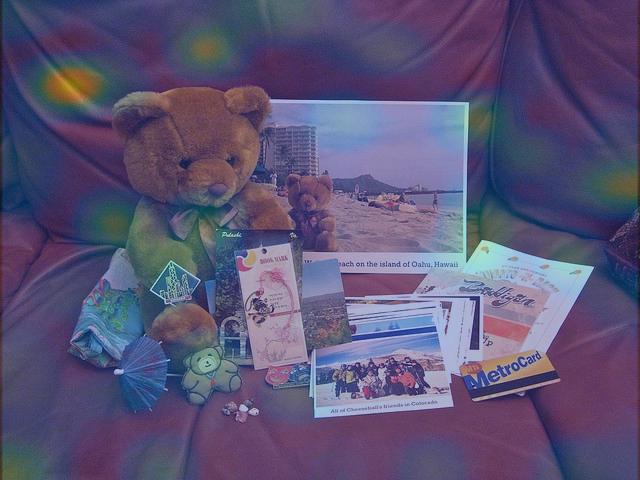}
&  \ \includegraphics[width=0.16\linewidth]{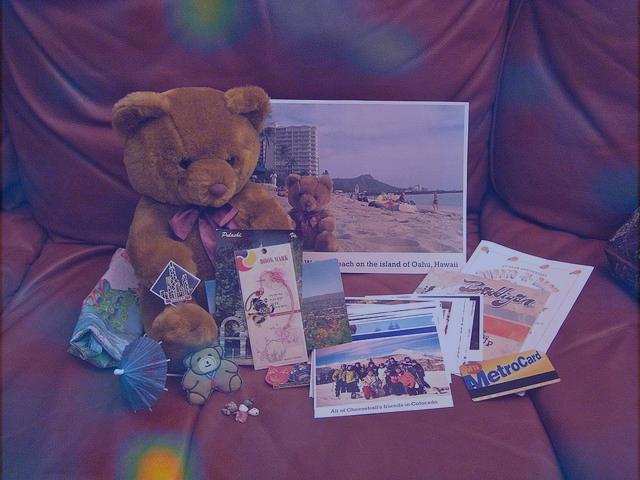} \\
% ------------- words ------------
what & is & the & bear & \textbf{sitting} & \textbf{on} \\
\end{tabular}
\end{center}
\vspace{-0.7cm}
\caption{Dynamic attention maps generated by the proposed model. Input answer: `\emph{Bowtie}' (top), `\emph{Couch}' (bottom). Because the conditioning answer is different the model generates totally different attention maps in producing the output question.}. \label{fig:att_bear}
\end{figure*}

\subsection{Results}
\keypoint{Overall} In the first experiment we report the overall iVQA performance on the test split. The results are shown in Table~\ref{tab:main_table} with both the standard linguistic metrics, as well as our ranking accuracy metric. From the table, we can make the following observations: (i) Unlike VQA the margin between the no-image case (A), and the full model (Ours) is dramatic. The ranking accuracies are more than doubled, and the language metrics show similarly striking improvements. This demonstrates that unlike conventional VQA \cite{VQA2.0}, the `V' does matter in iVQA. \
(ii) The margin between the image + answer type (I+AT) and image-only (I) setting exists, but is less significant. This shows that while it is a useful hint  for an iVQA model to know the question type, it still really needs the actual semantic answer to generate the right questions. E.g., rather than just knowing that it was counting something (answer type), the model does need to know \emph{how many objects were counted} (answer) in order to generate the right question specifying what object type needs to be counted -- as there may be other objects that could be counted.  
(iii) The margins between  (I) and (I+AT)  and the full model are also striking. This demonstrates that as a test of multi-modal intelligence, iVQA reassuringly requires both modalities in order to do well. 
(iv) VQG+VQA indeed performs better than the vanilla VQG model by making the generated question more answer conditional, but it is still weaker than the captioning adapted models (NN and SAT) or our proposed model. The reason is that \fnote{the VQG model has too low {\em sensitivity}, thus is unable to provide the right question candidates for  VQA model to select. Also, VQA model can be easily fooled by similar questions not relevant to the image because  VQA models are trained to distinguish different answers rather than questions.}
%due to the language bias, it is difficult for the VQA model to identify the right question from multiple candidates where the same answer applies.}
(v) The  captioning adapted models (NN and SAT) perform well, but are still inferior to the proposed model which is specifically designed for iVQA.

\vspace{0.1cm}\keypoint{Human study} The human study is applied on a subset of 1000 samples, and evaluates the models in a way that is robust to open-ended question generation. Results in Table~\ref{tab:main_table} show that (i) our model performs the best among all competitors; (ii) the human study scores are highly correlated with the proposed ranking metric. Specifically, the Pearson correlation coefficient between manually labelled scores and the proposed acc@1 and acc@3 metrics are 0.917 and 0.981 separately, while the best performing linguistic measure (CIDEr) can only reach 0.898, which demonstrates the effectiveness of the proposed ranking metric. Since human evaluation is expensive, the proposed ranking metric is a reasonable and cost-effective alternative.
%The PCC of all metrics are listed in the supplementary material.}  

\vspace{0.1cm}\keypoint{Qualitative Results} Examples of questions generated by  our model are shown in Fig.~\ref{fig:qualitative}. The results illustrate a few interesting points: (i) The generated questions are highly conditional on both images and answers. Particularly, the same answers generate different questions for different images, unlike the situation in VQA \cite{vqa_analysis1}; and the same images generate very different questions when paired with different answers, showing  richer reasoning than in VQG \cite{vqg}. (ii) Unlike VQA, there are multiple reasonable questions that correspond to one image-answer pair. This is both due to alternative phrasing of the same question (`\emph{where is the bear?}','\emph{where is the teddy bear?}',`\emph{what is the teddy bear sitting on?}'), as well as multiple semantically distinct questions having the same answer (e.g., `\emph{are the children eating?}',`\emph{is the child wearing a shirt?}',`\emph{is the child wearing a hat?}'). Since the annotation is not exhaustive, the standard linguistic metrics could be misleading: the generated questions can be correct but  just have never been asked by the annotators. Our proposed ranking metric is more robust to this, as models are only scored according to how plausibly they rate the true question, rather than whether their open-world estimate of the question matches  annotated ground-truth.  Our human study evaluates the methods in a way that credits open-world question generation.  The open-ended question generation formulation of our model means it is straightforward to \emph{sample} the  distribution over questions given images and answers, in order to explore the model's beliefs.

\vspace{-0.3cm}
\subsection{Further analysis}
\vspace{-0.2cm}
\keypoint {Analysis by Failure Type}
The proposed  new evaluation metric enables us  to understand the mistakes each model makes. The results in Fig.~\ref{fig:distractorTypes} show the Rank-1 predictions of each model (highest scoring question) broken down according to the category of that prediction. It again shows the much superior performance of our iVQA model (I+A+Att+I$_s$): 32.67\% of the top ranked prediction is correct, which almost doubles that of the VQA model (16.4\%). But the main objective here is to analyse which distractor types are mistakenly ranked high: 
(i) Without being able to condition on the answers, the image-only  method (I) makes predictions dominated by contrastive (CT) distractor questions taken from similar looking images. 
(ii) The answer-only (A) and image+answer type (I+AT) methods make predictions that are dominated by distractor questions of the popular (PP) type. This suggests that these models are trying to rely on unconditional distribution of label statistics in order to solve iVQA. 
(iii) The VQA-based baseline instead is dominated by answer-related (RN) distractor. This suggests that the VQA model fails to correctly take into account image context (the `V' is not being accounted for \cite{vqa_analysis1,VQA2.0}), and is simply picking  questions that generate the corresponding answer independently of the required conditioning  on  image context. 
(iv) The nearest neighbour (NN) approach performs well and has an evenly distributed set of error types, but it is weaker than the proposed in correctly capturing the visual and answer conditions. It is reflected on the larger portion of CT and RN errors. 
(v) Our full model (I+A+Att+I$_s$) has the largest fraction of correct predictions and manage to suppress the plausible (PS) and contrastive (CT) errors. It still makes some mistakes of ranking the popular (PP) questions at the top, but still fewer than answer (A) and image+answer type (I+AT) alternatives.

\vspace{0.1cm}\keypoint{Ablation study}
We evaluate  the contributions of our key technical components:  dynamic attention, and multi-modal inference with both local and global semantic image features. The results  in Table \ref{tab:ablation} verify that each of these  components contributes to the final result, and the biggest contribution comes from the proposed dynamic attention model.

\begin{table}[t]
\footnotesize
\begin{center}
\begin{tabular}{@{} l|cc|cc @{}}
\toprule
 	& CIDEr	& BLEU-4	& Acc@1	& Acc@3 \\ 
\midrule 
I+A	& 1.541	& 0.196	& 29.929	& 48.356 \\ 
I+Att+A	& 1.698	& 0.208	& 32.547	& 51.024 \\ 
I+Att+A+I$_s$	& {\bf 1.750}	&   {\bf 0.214}	&  {\bf 33.636} &  {\bf 52.344} \\ 
\bottomrule
\end{tabular}
\end{center}
\vspace{-0.5cm}
\caption{Ablation study on the contributions of key model components. The results are obtained on the validation set.}
\label{tab:ablation}
\end{table}

\vspace{0.1cm} \keypoint{How dynamic attention helps} To illustrate  our dynamic attention in iVQA, we visualise   attention maps computed   during question generation  in Fig.~\ref{fig:att_bear}. From the examples,  we  see   that   focus of attention varies over time in an appropriate way according to the partially generated question as well as the answer. For example, Fig.~\ref{fig:att_bear}  shows that the model  focuses on  regions on and below the bear at the point of generating  the words `\emph{on}' and `\emph{wearing}' respectively. These examples also demonstrate that the learned iVQA model has achieved some degree of multi-modal visuo-linguistic understanding and  implicit reasoning capability.  

\vspace{0.1cm}\keypoint{Can iVQA help VQA?}
The problem of iVQA can be seen as a dual problem of VQA, so we investigate whether VQA performance can be boosted by applying iVQA to obtain a second opinion. Specifically, we conduct the experiments on the more challenging VQA 2.0 dataset and MLB-att \cite{mlb} trained on the training set is employed as the VQA model. During testing, the top 3 answers with the largest VQA score are served as answer candidates, then a second score is computed from the iVQA model. They are further combined by a score fusion network, whose output is utilised as the confidence of final prediction. Before the  VQA-iVQA fusion, the VQA model alone can achieve a validation accuracy of 57.85, while after the final model reaches an accuracy of 58.86, where the performance gain is mainly from the challenging number type (improved from 34.94 to 38.71). These results thus show that  iVQA  can indeed assist VQA. Actually iVQA can also be used as a diagnosis tool to extract the belief set of an existing VQA model, which is part of the ongoing work.

\keypoint{Contrasting VQA and iVQA as benchmarks} Finally, we discuss iVQA's interest as a benchmark compared to the conventional VQA. Two of the main kinds of bias that a VQA/iVQA model could use to cheat the benchmark are the output \emph{Prior bias}   (Ignore both inputs and predict only the most likely answer on VQA; use question frequency in iVQA), and \emph{Language bias} (ignore the image and use only the input -- question for VQA, answer for iVQA -- to predict the output). A good multimodal intelligence benchmark should require understanding and mutual grounding of both modalities, and should be hard to game by exploiting those biases. To analyses these issues we compare performance on iVQA and VQA benchmarks using Prior-alone and Language-alone (LSTM Q for VQA and Answer only for iVQA) baselines versus the full multi-modal model in each case (DeeperLSTM+Norm I for VQA~\cite{vqa_dataset}, and I+A model for iVQA).
The results in Table \ref{tab:vqa_ivqa} show that for VQA the bias-based baselines approach the performance of a full multi-modal model much more closely than the corresponding baselines do for iVQA. This suggests that VQA is easier to `game' (achieve an apparently high score without any image understanding or multimodal grounding), compared to iVQA. Thus we propose that iVQA makes a distinct and interesting benchmark for multimodal intelligence.

\begin{table}[htbp]
\footnotesize
\begin{center}
\begin{tabular}{@{} l|c|ccc @{}}
\toprule
 	& split	& Prior	& Language	& Language+Visual \\ 
\midrule 
iVQA (acc@1)	& test	& 3.94	& 14.59	& 28.44 \\ 
VQA (accuracy)	& test-dev	& 29.66	& 48.76	& 57.75 \\ 
\bottomrule
\end{tabular}
\vspace{-0.3cm}\caption{VQA vs. iVQA on bias-based gameability.}\label{tab:vqa_ivqa}
\end{center}
\end{table}

\vspace{-0.5cm}
\section{Conclusion}
We have introduced the novel task of inverse VQA as an alternative multi-modal visual intelligence challenge to the popular VQA paradigm. The analyses suggest that iVQA is appealing in terms of being less game-able via exploiting label-bias, more clearly requiring the mutual grounding and understanding of both visual and linguistic modalities, and naturally providing an open-world prediction setting. % In the future we will investigate jointly solving VQA and iVQA  in terms of coherent focus of attention, and question/answer predictions.

\vspace{0.1cm}
\noindent\textbf{Acknowledgements:}\quad This project received support from Natural Science Foundation of China (NSFC) grant \#61773117, \#61473086, \#61520106009, \#61533008,  and \#U1713209.

{\small
\bibliographystyle{ieee}
\bibliography{lstm}
}

\end{document}